\documentclass{article}
\usepackage{microtype}
\usepackage{graphicx}
\usepackage{subcaption}
\usepackage{booktabs}
\usepackage{tabularx}
\usepackage{amssymb} 
\usepackage{tcolorbox}
\usepackage[hyphens]{url}
\usepackage{hyperref}

\usepackage[accepted]{icml2026}

\usepackage{amsmath}
\usepackage{amssymb}
\usepackage{mathtools}
\usepackage{amsthm}
\usepackage[capitalize,noabbrev]{cleveref}
\usepackage[textsize=tiny]{todonotes}

\icmltitlerunning{Position: AI/ML Deepfake Research is Misaligned with AI-Generated Non-Consensual Intimate Imagery (AIG-NCII)}

\begin{document}

\twocolumn[
  \icmltitle{Position: AI/ML Deepfake Research is Misaligned with\\AI-Generated Non-Consensual Intimate Imagery (AIG-NCII)}

  \icmlsetsymbol{equal}{*}

  \begin{icmlauthorlist}
    \icmlauthor{Li Qiwei}{yyy}
    \icmlauthor{Wells Lucas Santo}{yyy}
    \icmlauthor{Sarita Schoenebeck}{yyy}
    \icmlauthor{Eric Gilbert}{yyy}
    
  \end{icmlauthorlist}

  \icmlaffiliation{yyy}{School of Information, University of Michigan, Ann Arbor, USA}

  \icmlcorrespondingauthor{Li Qiwei}{rrll@umich.edu}

  \icmlkeywords{deepfakes, non-consensual intimate imagery, dignity, misinformation, AI-generated media, AI NCII, subject-centric harms}

  \vskip 0.3in
]

\printAffiliationsAndNotice{}

\begin{abstract}
AI-generated non-consensual intimate imagery (AIG-NCII) is not adequately addressed in AI/ML literature regarding AI-generated media, commonly referred to as ``deepfakes''. While research on deepfakes currently focuses on its epistemic harms---or harms relating to truth and authenticity---this is misaligned with the dominant reality of generative AI abuse involving sexualized imagery. We conduct a landscape analysis of highly-cited works to demonstrate that technical interventions addressing deepfakes almost entirely ignore AIG-NCII, limiting the research ecosystem to authenticity detection tools. In this position paper, we argue that existing interventions address \textit{viewer-centric} epistemic harms, such as fraud or scams, but ignore \textit{subject-centric} dignity harms, such as AIG-NCII. We illustrate that knowing an image is synthetic does not mitigate harms to subjects and may, in some cases, even exacerbate them. We conclude by offering recommendations to realign the field, including updating threat models to consider subject-centric harms and addressing AIG-NCII in AI safety research. Finally, we caution that researchers should only engage in this high-risk domain if they implement safety guardrails for both subjects and researchers and establish partnerships with domain experts in sexual violence prevention. 

\begin{tcolorbox}[colframe=black, colback=white, center title]
\textbf{Content warning: }This paper includes content about online sexual violence and suicide.
\end{tcolorbox}

\end{abstract}

\section{Introduction}

The creation of non-consensual intimate imagery (NCII) is a concerningly dominant use case of generative AI, whose harms are still inadequately addressed. In a report released in January 2026, AI Forensics found that the majority use case of the commercial AI system ``Grok'' is undressing people without their consent~\cite{bouchaudGrok2026}. New York Times and the Center for Countering Digital Hate separately found that after Elon Musk shared an image of an undressed woman superimposed on a SpaceX rocket produced by Grok in late December, Grok was asked to generate 4.4 million images the following week, compared to roughly 311,000 the week before Musk's post~\cite{congerMusks2026}. It is estimated that more than 3 million of these images were sexualized, with at least 23,000 being of children~\cite{centerforcounteringdigitalhateGrok2026}. Yet, despite the empirical reality that sexual abuse is a primary driver of generative AI usage, and the growing development of applications with the ability to produce AI-generated non-consensual intimate imagery (AIG-NCII), the phenomenon is largely absent from extant research involving generative AI imagery. Instead, the technical interventions developed by the AI/ML community are largely designed for a different set of needs focused on threats to truth and trust. 

While current interventions are designed to address the question of whether a piece of media is authentic or synthetic, the inattention to AIG-NCII has resulted in oversights in how safety tools are developed. Drawing on the fundamental human right of dignity, as outlined in the Universal Declaration of Human Rights from the~\citet{UN}, we argue that interventions should also focus on addressing dignity harms that cause injury to the subject.

\textbf{This position paper argues that there is a structural misalignment between AI/ML research agendas and the reality of AI-generated media, with existing concerns focusing on viewer-centric epistemic harms, which ignore or even exacerbate subject-centric dignity harms.} Our contributions in this position paper are as follows:

\begin{enumerate}

\item We surface a disconnect between research motivation and the actual harms of deepfakes. Through a systematic landscape analysis of highly-cited works in top-tier venues between 2020 and 2025, we find scant consideration for cases of AIG-NCII, despite it likely accounting for more than half of generative AI usage~\cite{bouchaudGrok2026, centerforcounteringdigitalhateGrok2026, securityhero20232023}.

\item We analyze how this lack of engagement has resulted in the proliferation of ``authenticity'' tools. We demonstrate how these tools are insufficient for AIG-NCII and, in specific deployment contexts, could exacerbate subject-centric dignity harms.

\item Finally, we offer a series of recommendations for researchers and practitioners to realign technical interventions to account for AIG-NCII. We stress that all researchers who work in this domain must partner with domain experts in online sexual violence and utilize threat models that consider subject-centric dignity harms in technical interventions. 

\end{enumerate}

\section{The absence of AIG-NCII in harm reduction research}

A ``deepfake'' is a colloquial term used to describe AI-generated or altered content~\cite{dielHarm2025}. This includes deceptive political videos such as that of using the likeness of President Biden to tell voters not to vote in the New Hampshire primary during the 2024 election~\cite{bondHow2024}, or fraud content such as using the likeness of Elon Musk to offer an investment opportunity that led to billions of dollars in monetary losses~\cite{briannewDeepfakes2024}. A body of technical research has been developed to address deepfakes. As we show in our analysis, the vast majority of this work has been conducted without acknowledging harms of a sexual nature. While not all deepfake content is categorized as AIG-NCII, it still exists as a dominant form of deepfake, with reports estimating that up to 98\% of deepfake videos are pornographic in nature~\cite{securityhero20232023}. In fact, the term ``deepfake'' originates from a direct reference to sexual harm, with the term being derived from the username of a Reddit user who shared custom-made AI-generated videos depicting actresses performing sexual acts~\cite{burkell2019nothing}. Since then, however, the word has entered the global lexicon to refer to a broad range of AI-generated content, including AI-generated misinformation, scams, and political images and videos.

\subsection{AIG-NCII as a distinct harm category}

We use the acronym AIG-NCII (AI-generated non-consensual intimate imagery) to refer to the phenomenon of sexualized deepfake content of a specific individual that exists without their consent. This can refer to synthetic content that is created with generative AI technology to ``nudify'' or ``undress'' a subject  without their explicit agreement~\cite{van2020verifying}\footnote{It is important to note here that AIG-NCII does not require the subject to be fully nude~\cite{batool2024expanding}. Recent examples of AIG-NCII have included content that has attempted to remove hijabs from women~\cite{tenbarge2026grok}.}. Other terms that describe this same phenomenon include ``AI-NCII'', ``AI-generated image-based sexual abuse''~\cite{henry2026s}, synthetic nonconsensual explicit imagery (SNCEI)~\cite{wei2025we}, and ``deepfake pornography''~\cite{FURIZAL2025101882}. AIG-NCII as a concept is derived from traditional non-consensual intimate imagery (NCII)\footnote{Non-consensual intimate imagery (NCII) is also colloquially known as ``revenge pornography''.}. AIG-NCII is highly gendered, with the vast majority of those impacted being women and girls~\cite{securityhero20232023,bouchaudGrok2026}. At a societal level, AIG-NCII represents attempts to silence, de-platform, and de-legitimize agency both online and offline~\cite{maddocks2020deepfake}.

\textbf{From GANs to diffusion models.} The technical barriers to creating AIG-NCII have lowered precipitously over the last decade. Initially, from approximately 2017 to 2022, face-swapping was the primary mechanism used to create AIG-NCII, by superimposing a face onto another person's body. This was enabled by autoencoder architectures, popularized by open-source repositories like DeepFaceLab~\cite{perov2020deepfacelab}. The infamous Deepnude application, which ``undressed'' women in images, relied directly on the Pix2Pix conditional GAN architecture~\cite{isola2017image, wang2018high}. The current phase of AIG-NCII creation is driven by diffusion-based synthesis, enabled by the release of open-weight latent diffusion models such as Stable Diffusion~\cite{rombach2022high,schuhmann2022laion}. Unlike face-swapping, diffusion models allow for the generation of sexualized imagery via text-to-image prompting. Techniques such as Dreambooth~\cite{ruiz2023dreambooth} and Low-Rank Adaptation (LoRA)~\cite{hu2022lora} are now standard tools in AIG-NCII communities to fine-tune a specific individual's likeness with few reference photos. 

\textbf{Academic research directly contributes to AIG-NCII.}~\citet{han2025characterizing} found in an analysis of the online community for Mr. DeepFakes, a prominent marketplace for deepfake content, that there was a ``significant sharing of academic work'' on its website, with direct references to GitHub repositories for deepfake tools that cite 43 academic papers. Many of these deepfake tools are forked from open-source models directly from research papers, and many of these applications are simply wrappers around open-source research code~\cite{han2025characterizing, gibson2025analyzing}. Additionally, the use of nude bodies as training data, often collected without consent, gives rise to these capabilities~\cite{cintaqia2025stop}.
%from venues such as ICML

\textbf{Limitations of law and policy. }Despite legal prohibitions in the U.S. and abroad, enforcement remains insufficient, costly, and reactive, often placing the burden of discovery on the victim~\cite{sen_durbin_s3696_2024, qiwei2025law,noauthor_s146_nodate}. For example, while the U.K. Online Safety Act criminalizes sharing AIG-NCII, abuse has merely shifted to non-compliant platforms and encrypted apps like Telegram~\cite{telegram_2024}. Similarly, South Korea responded to major scandals~\cite{noauthor_korea_nodate,noauthor_cho_2020} by criminalizing possession~\cite{noauthor_south_nodate}, yet deepfake generation tools remain accessible. Moderation efforts on individual platforms face a similar ``whac-a-mole'' dynamic~\cite{ding2026stop}. When CivitAI banned real-person likeness models~\cite{noauthor_policy_nodate,maiberg__a16z-backed_2025,wagner2025perpetuating}, the models simply migrated to HuggingFace~\cite{maiberg__hugging_2025}.

\subsection{Landscape analysis of existing literature}

To quantify the misalignment between existing research concerns and the reality of AIG-NCII, we conducted an analysis of technical defense papers published between 2020 and 2025. Our analysis reveals that the literature addressing deepfake harms almost entirely ignores AIG-NCII.

\textbf{Methodology.} Our goal was to locate works that aimed to address harms regarding deepfakes or synthetic media more broadly. We queried Google Scholar for papers containing a specific set of keywords, using the following query: \textit{(``detection'' OR ``detector'' OR ``forensics'' OR ``recognition'' OR ``watermark'') AND (``deepfake'' OR ``synthetic image'' OR ``fake image'' OR ``diffusion'')}. This initial search criteria yielded 965 papers. Next, we filtered our results to papers published only at the top-tier venues: \textit{``CVPR'' OR ``ICCV'' OR ``ECCV'' OR ``NeurIPS'' OR ``ICML'' OR ``ICLR''}. We excluded workshop papers and included arXiv preprints that were returned in our search. Papers with more than 80 citations from other venues were also included. This brought our resulting dataset to 379 papers. Finally, we filtered the results down to the top 100 most cited papers, and manually excluded papers where diffusion models were utilized for unrelated computer vision tasks such as detecting tumors, detecting cars, or detecting cracks in steel. We also excluded one retracted paper. This process left a final dataset of 39 papers that we qualitatively analyzed. We examined each paper for engagement with AIG-NCII, with particular attention to the usage of the following terms: \textit{``non-consensual intimate imagery'' ``NCII'', ``revenge porn'',``sexual violence'' ``porn'', ``nudity'', ``undress'',``obscene''}. See Table 2 in the Appendix for the final list of 39 papers. 

\textbf{Results.} We categorized the 39 papers into three tiers of engagement with AIG-NCII.
\begin{enumerate}
\item \textbf{No mention (34 papers):} The paper frames the problem exclusively as misinformation, fraud, or technical artifact detection.
\item \textbf{Mention only (5 papers):} The authors reference AIG-NCII terms in passing, such as in the introduction or broad impact statement, but the proposed technical method remains generic.
\item \textbf{Technical implementation (0 papers):} The authors design an intervention with a threat model specific to AIG-NCII.
\end{enumerate}

Our analysis affirms existing research stating that deepfake research is overwhelmingly motivated by harms related to trust, fraud, and political misinformation ~\cite{rini2020deepfakes, harrisVideo2021}. 

\begin{figure}
    \centering
    \includegraphics[width=0.9\linewidth]{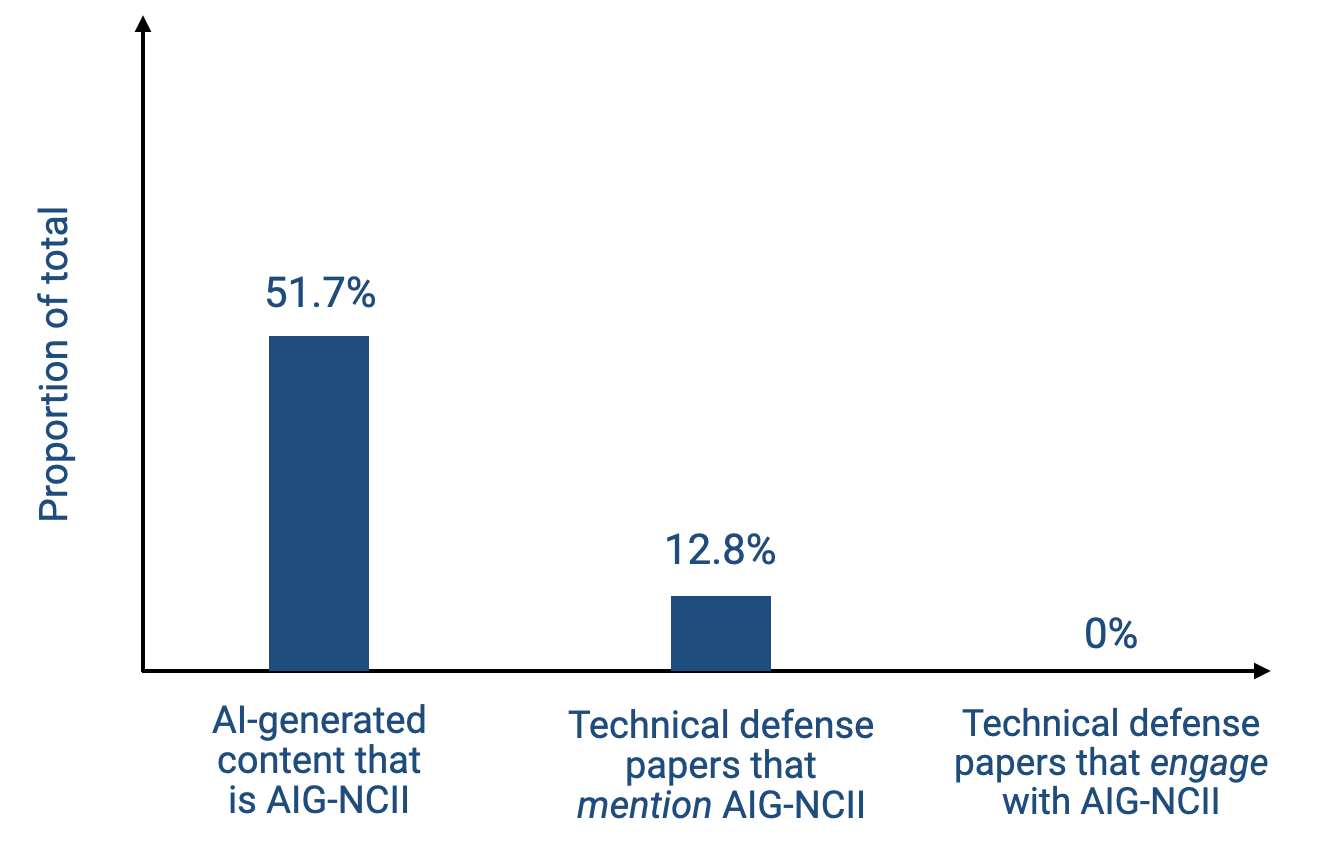}
    \caption{Proportion of AI-generated content that is classified as AIG-NCII~\cite{bouchaudGrok2026} compared to the proportion of technical defense papers that merely mention AIG-NCII. No papers found in our landscape analysis meaningfully engaged with AIG-NCII.}
    \label{fig:comparison}
\end{figure}

\section{Review of authenticity-based interventions} 

In order to deal with harms pertaining to truth, our analysis found that the AI/ML community has coalesced around a series of tools designed to distinguish \textit{synthetic} from \textit{authentic} media. We categorize these technical interventions into three primary paradigms: detection, provenance, and watermarking. While they differ significantly in their implementation, they share one foundational assumption, that \textit{truth-verification, or authenticity, is the primary proxy for safety.} In this section, we review these three paradigms and note the assumptions that underpin their design.

\subsection{Detection}

Current AI detection attempts to approximate a classification problem, wherein a model learns a decision boundary between the distribution of authentic media and synthetic media. The primary assumption of this paradigm is that a robust decision boundary exists, and that successfully determining the boundary between authentic and synthetic is a sufficient condition for addressing harm.  

While earlier literature focused on identifying GAN-specific artifacts in the media asset~\cite{frank2020leveraging}, the proliferation of diffusion models has required a fundamental shift in the feature extraction process for identifying synthetic content. More recent detection methods target the unique fingerprints introduced by the iterative de-noising process of diffusion models. For example, DIRE~\cite{wang2023dire} utilizes the observation that diffusion-generated synthetic images show lower reconstruction error inverted through a pre-trained diffusion process as compared to authentic images. Similarly,~\citet{corvi2023detection} identified and analyzed distinct spectral traces left by the Gaussian noise scheduling that are inherent to latent diffusion models. To address the rapid evolution of generator architectures (e.g., Stable Diffusion~\cite{rombach2022high} and FLUX~\cite{blackforestlabs_flux1tools_2024}), research has increasingly moved utilizing feature spaces from foundational vision-language models like CLIP~\cite{radford2021learning} to identify synthetic semantic patterns that generalize across architectures~\cite{ojha2023towards}. 

\subsection{Provenance}

Provenance methods, as defined by the~\citet{c2pa_specification_2.3} technical specification, attempt to establish a history of modifications for a media asset. Unlike methods in detection, this paradigm for protection relies on cryptographically verifiable information that can be used to verify that an asset is free from tampering. At each point of asset creation or modification, a digital signature is bound to a hash of the pixel data of the asset, along with a manifest containing metadata assertions, such as content ownership and timestamp~\cite{rosenthol2022c2pa}. The assumption guiding provenance methods is that having a verifiable chain-of-custody resolves trust in where a piece of media originated, and whether it was edited along the way. In other words, interventions following the provenance paradigm are concerned with tracking the lineage from a source asset to any of its modified outputs, such that the lineage itself is an indicator of authenticity.

\subsection{Watermarking}
Content watermarking techniques aim to embed signals invisible to the human eye directly onto the media content at generation, to signal that content as synthetic. Unlike metadata, which can be easily stripped and removed from the asset, watermarks aim to be more robust against being removed, even with cropping, filters, and other image transformations. Approaches within this paradigm include latent watermarking~\cite{fernandez2023stable} and sampling-based watermarking~\cite{wen2023tree}. Industry implementations, such as Google's SynthID~\cite{deepmind_synthid_2025}, embeds signals onto the media that are detectable only when paired with a specialized detector model or decoding mechanism. Watermarking is often used to help with post-hoc detection.

\section{Consequences of ignoring AIG-NCII}

The intervention paradigms around detection, provenance, and watermarking are meant to restore the viewer's ability to discern what is authentic. For AIG-NCII cases, however, the subject suffers from violation of their dignity regardless of authenticity. In this section, we discuss possible consequences overlooking these dignity violations.

\subsection{Neglect of subject-centric dignity harms}

\begin{figure}
    \centering
    \includegraphics[width=0.75\linewidth]{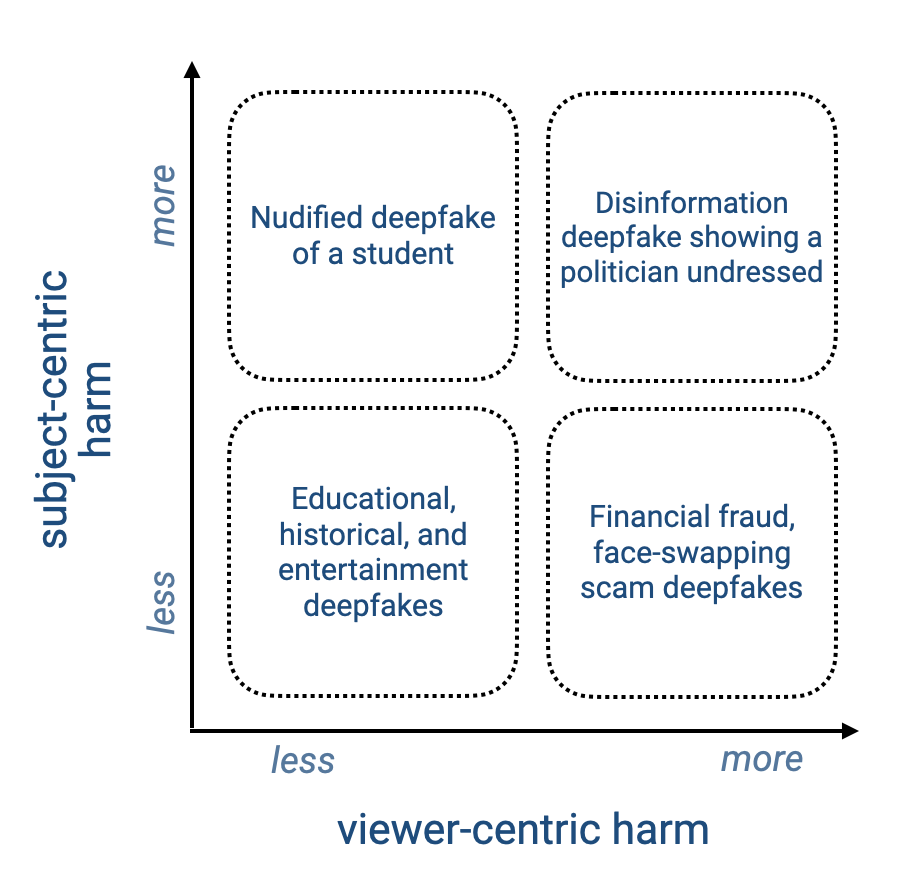}
    \caption{Examples of deepfakes that cause harm to the viewer, subject, or both parties.}
    \label{fig:viewer-subject}
\end{figure}

The exclusion of AIG-NCII in deepfake harm reduction research has resulted in an ecosystem that solely focuses on epistemic harms, or harms relating to truth and authenticity. However, as~\citet{harrisVideo2021} notes, these concerns often overlook the severity of harms to subjects. Following~\citet{chesney2019deep} and~\citet{citron2018sexual} and drawing from the UN Universal Declaration of Human Rights~\cite{UN}, we argue that AIG-NCII violates the fundamental human right of dignity for deepfaked subjects, which is distinct from epistemic harms to viewers. We build upon the framework by~\citet{olteanu2025ai} to distinguish between harms to the viewer and the subject:

\textbf{Viewer-centric harms} encompass both individual deception and broader societal epistemic degradation. On an individual level, this includes fraud, such as scams mimicking a friend's likeness. However, as~\citet{rini2020deepfakes} argues, this actually presents a societal harm due to the erosion of authenticity and the destabilization of shared reality~\cite{chesney2019deep}. If one was led to believe false information, they may have suffered a viewer-centric epistemic harm.

\textbf{Subject-centric harms} occur when a person's likeness is utilized without consent, regardless of whether the viewer is deceived. Drawing on~\citet{citron2018sexual}'s idea of sexual privacy, the harm arises from the undignified presentation of the subject and the resulting loss of autonomy. At a high level, a dignity harm occurs when one's likeness is used in ways they did not \textit{consent} to. We highlight two instances of this harm that map onto specific AI/ML techniques. The first is non-consensual identity preservation, in which a model retains a recognizable likeness of a specific person without their consent. Defenses against this mode may aim to prevent identity retention---for example, by disrupting a model's ability to learn or reproduce a specific subject's features~\cite{van2023anti}. The second is non-consensual modification, in which the subject's likeness is intentionally preserved, but the image is altered in ways they did not consent to, such as removing clothing or face-swapping with pornographic videos.

Viewer-centric and subject-centric harms are orthogonal, as shown in Figure \ref{fig:viewer-subject}. While these harms often overlap---for instance, a political deepfake may deceive constituents (viewer-centric) while damaging the politician's reputation (subject-centric)---the mechanisms required to address them differ. The lack of engagement with subject-centric harms has led to a lapse in defense design, where merely labeling content as ``synthetic'' does not mitigate subject-centric harms. The misalignment between technical capabilities and safety needs is so acute that even platform governance bodies have begun questioning use of these tools. The Oversight Board for Meta recently noted that ``labeling manipulated content is not appropriate in this instance because the harms stem from the sharing and viewing of these images---and not solely from misleading people about their authenticity''~\cite{oversightboard_deepfake_intimate_2024}. Indeed, authenticity-based interventions do not mitigate harms to subjects, especially given that prominent online forums that host AIG-NCII already routinely label the content as fake~\cite{han2025characterizing}. 

\subsection{Authentic $\neq$ safe}

The current trajectory of deepfake defense research optimizes for authenticity metrics. When applied to AIG-NCII, this creates a fundamental category error, treating authenticity as a proxy for safety. This substitution fails because, unlike political misinformation where falsehood is the primary harm, sexual violence is defined by the absence of consent, which is violated regardless of whether an image is authentic or synthetic. As illustrated in Table \ref{tab:orthogonality}, the axis of artificiality (what existing interventions measure) is orthogonal to the axis of consent (what determines safety). Relying on authenticity-based tools creates a blunt instrument that cannot address the case of AIG-NCII, because it conflates non-consensual synthetic imagery with that of consensual synthetic imagery. At the same time, the ecosystem risks building over-censoring tools that stifle legitimate expression while failing to address traditional non-consensual imagery simply because it is ``authentic''. We argue that until technical interventions can account for this orthogonality, evidence of authenticity should not be treated as sufficient evidence of safety.

\begin{table}[t]
    \centering
    \small % Reduces font size to make it fit better
    \renewcommand{\arraystretch}{1.4} 
    \begin{tabularx}{\columnwidth}{@{}l|X|X@{}} % X columns auto-wrap text
         & \textbf{Safe} & \textbf{Harmful} \\ \hline
        \textbf{Synthetic} % Shortened label to save space
        & Self-expression of artistic nudity using AI
        & AIG-NCII \\ \hline
        \textbf{Authentic} % Shortened label to save space
        & Consensual \mbox{pornography}
        & Traditional NCII \\ 
    \end{tabularx}
    \vspace{0.5cm}
    \caption{The orthogonality of harm. The current detection paradigm can only distinguish between synthetic and authentic media. However, safety falls along an orthogonal axis that is determined by consent.}
    \label{tab:orthogonality}
\end{table}

\subsection{Potential misuse of authenticity tools}

The implicit assumption guiding current research agendas is that safety can be achieved once the most accurate authenticity model is developed. We challenge this assumption. In the following contexts, we demonstrate how authenticity interventions, when built without consideration for a subject-centric threat model, may actually exacerbate harms to subject dignity.

\paragraph{When used by online platforms.}

Current regulations, such as the EU AI Act~\cite{eu_ai_act_article50}, as well as platform policies on Meta~\cite{meta_ai_labeling_2024} and TikTok~\cite{tiktok_ai_generated_content}, prioritize the labeling of synthetic media that is identified. While effective for viewer-centric epistemic harms, this model of public labeling could backfire for AIG-NCII. A label stating that an image is made with AI does not address the fact that the image is shared on online platforms. While some platforms explicitly ban nudity, AIG-NCII also includes cases where the subject is not fully nude, which is ignored by these regulations~\cite{batool2024expanding}. When labeling content is prioritized over its removal, this could create a perverse outcome where the abuser may be protected from moderation consequences so long as they are transparent about the synthetic, AI-generated nature of the image.

\paragraph{When used by abusers.}

Research in online sexual violence indicates that perpetrators are often driven by an assertion of power over a victim, rather than by sexual gratification on its own~\cite{henry2024image,henry2016sexual}. In fact,~\citet{marini2024real} has shown that people are less aroused when they find that an image is identified as AI-generated. As noted by~\citet{massanari2017gamergate}, these communities are not passive consumers but active participants who ``demonstrate technological prowess'' in aggregating disparate pieces of content to target and verify the identity of victims. These same communities may use authenticity-based tools to locate content in order to further harass and dox victim-survivors. In this context, authenticity verification may become a tool that allows users to sort through authentic and synthetic imagery for abuse.

\paragraph{When used against NCII victim-survivors.}

Finally, the utility of authenticity labels is not consistent for victim-survivors, fluctuating depending on the harm that the subject is dealing with. For victim-survivors of traditional NCII, plausible deniability may actually offer a safety mechanism to protect their dignity. In this context, ambiguity offers a form of protective cover for victim-survivors. If a detection system definitively labels traditional NCII as ``authentic'', it inadvertently acts as a verification tool for abusers, confirming the victim's exposure to the public. By removing this uncertainty, technical interventions may out victims who could otherwise have maintained some degree of social safety by casting doubt on the image's veracity.

\section{Recommendations}

Addressing AIG-NCII is an extremely difficult task, and it is one that faces significant barriers both ethically and technically. Given the sensitive nature of this abuse, we cannot always propose definitive solutions. In this section, we raise recommendations to best address the problem. Ultimately, we believe there is a need for both technical and social pathways for addressing the proliferation of AIG-NCII, and this requires coordination between AI/ML researchers and practitioners, social scientists, policy makers, sexual violence prevention experts, and victim-survivor advocates.

\textbf{R1. Decouple epistemic and dignity harms.} As we have shown, care must be taken when applying authenticity markers. If a system identifies potential AIG-NCII, it should not apply a public label, as this keeps content visible, and may exacerbate harm. Instead, detection should serve as a backend flag that triggers precautionary handling (suppression or triage) treating AIG-NCII like traditional NCII when content depicts real people in sexualized contexts. We urge the research community to study the specific trade-offs of labeling AIG-NCII before deploying detection tools. Future work should identify how to verify content in a way that protects the privacy of victims while empowering deepfake subjects to defend themselves. Furthermore, we must weigh the asymmetrical harm of errors in labeling. While temporarily restricting consensual content is often reversible, failing to intervene in situations of sexual violence imposes irreversible harms. Future work must rigorously evaluate how transparency standards designed for misinformation may inadvertently endanger privacy.

\textbf{R2. Elevate subject-centric dignity harms.} The violation of dignity must be elevated to the same tier as political misinformation. Researchers should mirror the shift in computer security towards analyzing intimate partner violence (IPV), where the adversary is personally known to the victim-survivor~\cite{chatterjee2018spyware,havron2019clinical,freed2018stalker}. In AIG-NCII, the adversary is often an actor equipped with limited reference photos and parameter-efficient fine-tuning techniques. Under this framework, the goal shifts from maximizing detection accuracy to minimizing identity preservation. Defenses are successful if they prevent the reproduction of a specific identity or make it more challenging. Furthermore, we must reject the assumption that publicly available data is synonymous with consensual data. Privacy is violated by the migration of information outside its intended context~\cite{nissenbaum2004privacy}. While modeling general human features may be necessary in some contexts, the non-consensual ingestion of nude or sexualized bodies crosses an ethical red line~\cite{stark2018facial,scheuerman2021datasets}. Researchers should abandon the unethical curation of datasets from sensitive domains~\cite{cintaqia2025stop}.

\textbf{R3. Restrict high-risk research assets.}
We call for a re-evaluation of open-release norms for architectures explicitly optimized for high-fidelity identity retention and inpainting. While open science is a core value of the field, an emerging consensus in AI Safety literature recognizes that the risks of open-sourcing highly capable, dual-use models often outweigh the benefits~\cite{widder2022limits,solaiman2023gradient,seger2023open}. Models and fine-tuning techniques that demonstrate state-of-the-art performance (such as ``cloning'' the likeness of a person from a few photographs) should be subject to gated or researcher-only access. By placing additional friction on the tools of creation, we can reduce the size of the threat downstream. 

\textbf{R4. Consider proactive prevention.} The field should move beyond post-hoc detection and engage seriously with adversarial defense mechanisms against inpainting, one of the primary techniques for nudification. Adversarial immunization introduces human-imperceptible perturbations that disrupt internal representations in generative models, preventing style mimicry or inpainting~\cite{jeon2025advpaint,guo2025anti,van2023anti,shan2023glaze}. While these defenses can be brittle against transformations~\cite{honig2025adversarial}, recent work may be closing this gap by making perturbations harder to remove~\cite{kim2026blurguard} and dramatically lower the per-image protection cost~\cite{ozden2025diffvax}. Even imperfect defenses raise the cost for lay-abusers and disrupt the generative pipelines.

\textbf{R5. Develop safety-aligned metrics.} Success must be measured by harm reduction, not merely detection accuracy. We need metrics that verify whether systems actually reduce the prevalence of abuse. This requires creative evaluation strategies. For instance,~\citet{cretu2025evaluating} utilized ethical proxies to assess child-safety filters without generating actual CSAM. This may offer inspiration for evaluating AIG-NCII interventions without producing harmful content.

\textbf{R6. Integrate AIG-NCII into AI Safety.} Subject-centric harms must be elevated to a core concern within the definition of AI Safety. Scholars have increasingly critiqued the mainstream discourse for focusing on existential risks while excluding present-day harms~\cite{gyevnar2025ai,ahmed2023building,hazra2025ai}. We further note that the dominant model-side safety techniques (e.g., concept erasure and NSFW filtering~\cite{gandikota2023erasing,schramowski2023safe}) are insufficient for AIG-NCII because they assume a cooperative model operator. The AIG-NCII ecosystem is defined by open models and non-compliant platforms~\cite{gibson2025analyzing}. AIG-NCII therefore requires safety research that does not depend on operator goodwill. If the field is to meet its goal of protecting human welfare, the scope of AI Safety must expand to include the immediate violence of AIG-NCII. Conference organizers should recognize the scale of this abuse and allocate the same rigor and resources currently afforded to sub-fields such as algorithmic bias and toxic content generation~\cite{weidinger2021ethical}.

\textbf{R7. Establish ethical partnerships and guardrails.} Research into AIG-NCII poses unique ethical and psychological challenges and is not suitable for all researchers. Guardrails must be implemented to mitigate secondary traumatic stress for those who review sensitive content ~\cite{williamson2020secondary}. Crucially, technical researchers must avoid speculatively deriving threat models in isolation. Work must be empirically grounded and co-designed with domain experts in online sexual violence and victim advocacy~\cite{costanza2020design}. These experts should be integrated as partners during the initial design phase, rather than merely consulted for post-hoc validation.

\textbf{R8. Account for victim-survivor plurality.} Interventions must respect the spectrum of survivor needs. As noted by~\citet{mcglynn2019kaleidoscopic}, victim-survivors may prioritize drastically different outcomes, ranging from criminal prosecution to content removal. A one-size-fits-all technical solution cannot serve all these needs. Researchers should leverage frameworks of restorative justice to design flexible tools that respect survivor agency, rather than imposing a monolithic technical ``solution''~\cite{schoenebeck2021drawing}.

\textbf{R9. Acknowledge that social harms require social interventions for remediation.} While existing interventions are focused on detecting and identifying the synthetic nature of deepfakes, this does not actually resolve any of the dignity-based harms that have been inflicted on victim-survivors. Experts in interpersonal violence (IPV) and sexual violence reduction should be consulted when developing possible interventions to remediate the harms of AIG-NCII.

\section{Alternative Views}

In this section, we address the primary objections to our arguments that AI/ML research must account for AIG-NCII.

\textbf{AV1: Addressing social harms is the domain of law and policy, not AI/ML.}
One objection to this paper's argument is that AI/ML researchers are only responsible for optimizing technical capabilities, while the regulation of those capabilities belongs to policymakers. From this perspective, AIG-NCII is fundamentally a legal problem that arises from under-resourced legal systems and platforms that fail to enforce abuse. How generative AI tools are used is out of scope for AI/ML researchers. 

\textbf{Response:} We agree that law and policy are essential, but we reject the premise that research into technical protections is therefore irrelevant. First, the timing mismatch between AI development and legal enforcement makes relying solely on the law unfeasible. The legal system operates on timescales of years, and generative AI capabilities evolve in weeks. In the example of the UK Online Safety Act, by the time the legislation passed, the dominant mechanism for AIG-NCII had already shifted from face-swapping to diffusion synthesis. The decision to work on defense methods against AIG-NCII is as much a research choice as it is to work on other technical innovations such as LoRA. Much like how the fight against Child Sexual Abuse Material (CSAM) has involved legal, technical, and advocate coordination, we argue that technical protections can work in tandem with legal efforts against AIG-NCII, which requires a multi-pronged approach.

\textbf{AV2: Prevention of AIG-NCII is technically intractable.}
Even if researchers accept responsibility, proposed interventions simply do not work. Adversarial defenses such as inpainting perturbation are brittle and easily defeated by compression or model updates~\cite{sun2023critical,guo2025anti,goodfellow2014explaining,athalye2018obfuscated,honig2025adversarial}. Furthermore, the sheer scale of the internet makes protecting every user's photo impossible, and sophisticated abusers can easily bypass protections using slightly altered inputs. Therefore, proposing ``better defenses'' offers false hope to victim-survivors.

\textbf{Response:} We concede that no technical intervention will ever be a perfect shield. However, the goal of technical defense is not perfection, but friction. Currently, an abuser can generate AIG-NCII in minutes with little to no cost or technical skill. Adversarial defenses, even if imperfect, raise the cost of abuse by requiring knowledge to bypass. This friction may disrupt casual abuser (teenagers, ex-partners) who account for a significant volume of harassment but lack the sophistication to break these defenses. Additionally, while current defenses are brittle, this should be a call for further research, not abandonment. 

\textbf{AV3: The cultural and institutional incentives of AI/ML research makes engagement with AIG-NCII unrealistic.} Even if researchers accept the framing of subject-centric harms, there are structural barriers to working on such topics. Lack of institutional support, concerns about the ability to publish, and reviewer discomfort with sexualized topics makes it difficult for researchers to engage with AIG-NCII. 

\textbf{Response:} We acknowledge these incentive structures, but argue that the field cannot claim AIG-NCII is too peripheral to engage with when it is a direct downstream product of mainstream research decisions. The open datasets used to train widely-deployed diffusion models contained non-consensual images of human bodies, including CSAM~\cite{thiel2023identifying}. The techniques now used for AIG-NCII (e.g., face-swapping, inpainting, fine-tuning on specific individuals) were each developed and celebrated at top venues. Having produced the capabilities, the field also bears responsibility for the mitigating their negative societal impacts. The field has shifted before. Algorithmic fairness was a fringe topic a decade ago. It took several key papers to point to the problem~\cite{buolamwini2018gender,angwin2022machine} and efforts to form dedicated conferences (FAccT and AIES, both in 2018). The transition required individual researchers willing to legitimize the topic, conference organizers willing to allocate space, and senior scholars willing to advise students working on it. None of these moves required the field to \textit{first} be comfortable. In fact, comfort followed the work, not the other way around. AIG-NCII can follow a similar trajectory if researchers refuse to accept ``taboo'' as a reason for inattention. 

\section{Conclusion}

In this position paper, we show that there is a fundamental misalignment between the current technical interventions for deepfakes that address viewer-centric epistemic harms and the prevailing reality of subject-centric dignity harms in the form of AIG-NCII, which account for the majority of generative AI usage~\cite{centerforcounteringdigitalhateGrok2026,bouchaudGrok2026,securityhero20232023,gibson2025analyzing}. We urge the AI/ML community to realign its priorities to address these harms, else we risk exacerbating harms to victims of AIG-NCII. At the same time, we offer our call to action in the form of recommendations with a necessary constraint. Research into protections against AIG-NCII should only be undertaken when adequate safety protocols are established, including mitigating harm for researchers and establishing substantive partnerships with victim-advocates and sexual violence prevention experts. Ultimately, the research community bears a responsibility to ensure that our definitions of AI safety protect not only truth, but also the dignity of people.

\section*{Acknowledgments}
We thank Amna Batool, Rosanna Bellini, and Su Lin Blodgett for conversations that helped refine the framing of this work. This material is based on works supported by the National Science Foundation under Grant 2311102.

\bibliography{example_paper}
\bibliographystyle{icml2026}

\newpage
\appendix
\onecolumn
\begin{table}[ht]
\centering
\renewcommand{\arraystretch}{1.2}
\begin{tabularx}{0.9\textwidth}{@{} X l c @{}}
\toprule
\textbf{Paper surveyed} & \textbf{Venue} & \textbf{Mentions AIG-NCII} \\ 
\midrule
\citet{frank2020leveraging} & ICML & \\
\citet{wang2023dire} & ICCV & \\
\citet{corvi2023detection} & ICASSP & \\
\citet{hsu2020deep} & Applied Sciences & $\checkmark$ \\
\citet{ahmadi2020redmark} & Expert Systems & \\
\citet{zhao2023recipe} & arXiv preprint & \\
\citet{raza2022novel} & Applied Sciences & $\checkmark$ \\
\citet{ricker2022towards} & arXiv preprint & \\
\citet{liu2024forgery} & CVPR & \\
\citet{patel2023improved} & IEEE Access & $\checkmark$ \\
\citet{yang2024gaussian} & CVPR & \\
\citet{ricker2024aeroblade} & CVPR & \\
\citet{bammey2023synthbuster} & IEEE OJSP & \\
\citet{chen2024drct} & ICML 2024 & \\
\citet{hamid2023improvised} & Springer IJSA & \\
\citet{luo2024lare} & CVPR 2024 & \\
\citet{liu2023watermarking} & arXiv preprint & \\
\citet{ma2023exposing} & arXiv preprint & \\
\citet{rahman2023artifact} & IEEE ICIP & \\
\citet{damer2023mordiff} & arXiv preprint & \\
\citet{yu2024diffforensics} & CVPR & \\
\citet{guarnera2024level} & Intell. Sys. Conf. & $\checkmark$ \\
\citet{wu2025generalizable} & arXiv preprint& \\
\citet{aghasanli2023interpretable} & ICCV & \\
\citet{song2023robustness} & arXiv preprint& \\
\citet{sun2024diffusionfake} & NeurIPS & \\
\citet{lei2024diffusetrace} & arXiv preprint & \\
\citet{sinitsa2024deep} & WACV & \\
\citet{song2024learning} & NeurIPS & \\
\citet{zhu2022pixelhop} & ICASSP & \\
\citet{zhang2022patch} & AAAI & \\
\citet{wen2025spot} & arXiv preprint & \\
\citet{zhang2024attack} & NeurIPS & \\
\citet{liao2021imperceptible} & arXiv preprint & $\checkmark$ \\
\citet{keita2025bi} & Expert Systems & \\
\citet{kang2025legion} & arXiv preprint& \\
\citet{zhang2023diffusion} & arXiv preprint& \\
\citet{sun2023critical} & ICML & \\
\citet{ivanovska2023face} & arXiv preprint & \\

\bottomrule
\end{tabularx}
\caption{39 papers surveyed, listed in order of number of citation each paper received in January 2026.}
\end{table}
\end{document}